\documentclass{article}
\usepackage{multirow}
\usepackage{booktabs}
\usepackage[utf8]{inputenc} 
\usepackage[T1]{fontenc}    
\usepackage{hyperref}       
\usepackage{url}            
\usepackage{booktabs}       
\usepackage{amsfonts}       
\usepackage{nicefrac}       
\usepackage{microtype}      

\usepackage{amsmath,amssymb,amsfonts,amsthm}
\usepackage{bm}
\usepackage{graphicx}

\usepackage{float}
\usepackage{algorithm}
\floatstyle{ruled}
\newfloat{algo}{tbp}{loa}
\providecommand{\algorithmname}{Algorithm}
\floatname{algo}{\protect\algorithmname}

\DeclareRobustCommand{\ov}[1][\qedsymbol]{%
	\ifmmode \mathqed
	\else
	\leavevmode\unskip\penalty9999 \hbox{}\nobreak\hfill
	\quad\hbox{$\blacksquare$}%
	\fi
}

\newcommand{\diag}{\mathrm{diag}}

\newcommand{\gFDR}{\mathrm{gFDR}}
\newcommand{\EQ}{\mathrm{EQ}}

\newcommand{\swap}{\mathrm{swap}}

\DeclareMathOperator{\ep}{E}
\DeclareMathOperator{\Cor}{Cor}

\DeclareMathOperator{\Cov}{Cov}

\newcommand{\argmin}{\mathop{\arg\min}}

\newcommand{\real}[1]{{\mathbb R}^{#1}}

\newcommand{\bA}{\mathbf A}

\newcommand{\bB}{\mathbf B}

\newcommand{\bD}{\mathbf D}

\newcommand{\bS}{\mathbf{S}}
\newcommand{\bs}{\mathbf{s}}

\newcommand{\bw}{{\mathbf w}}

\newcommand{\Xtilde}{\tilde{X}}
\newcommand{\bX}{\mathbf X}

\newcommand{\bXtilde}{\widetilde{\bX}}

\newcommand{\bz}{{\mathbf z}}
\newcommand{\bztilde}{\tilde{\bz}}

\newcommand{\greekbold}[1]{\mbox{\boldmath $#1$}}

\newcommand{\bbeta}{\greekbold{\beta}}

\newcommand{\eps}{\epsilon}

\newcommand{\bSigma}{\greekbold{\Sigma}}

\newcommand{\Scal}{{\mathcal S}}

\newcommand{\deq}{\overset{\text{d}} {=} }  

\usepackage{natbib}
\theoremstyle{definition} \newtheorem{definition}{Definition}
\theoremstyle{plain} \newtheorem{theorem}[definition]{Theorem}
\theoremstyle{plain}

\usepackage{authblk}

\begin{document}

\title{Deep-gKnock: nonlinear group-feature selection with deep neural network}

\author[1]{Guangyu Zhu}
\author[2]{Tingting Zhao}
\affil[1]{Department of Computer Science and Statistics, University of Rhode Island}
\affil[2]{Department of Electrical  \& Computer Engineering, Northeastern University }


	\maketitle
	
	\begin{abstract}
		Feature selection is central to contemporary high-dimensional data analysis. Grouping structure among features arises naturally in various scientific problems. Many methods have been proposed to incorporate the grouping structure information into feature selection. However, these methods are normally restricted to a linear regression setting. To relax the linear constraint, we combine the deep neural networks (DNNs) with the recent Knockoffs technique, which has been successful in an individual feature selection context. We propose Deep-gKnock (Deep group-feature selection using Knockoffs) as a methodology for model interpretation and dimension reduction. Deep-gKnock performs \emph{model-free} group-feature selection by controlling group-wise False Discovery Rate (gFDR). Our method improves the interpretability and reproducibility of DNNs. Experimental results on both synthetic and real data demonstrate that our method achieves superior power and accurate gFDR control compared with state-of-the-art methods.
	\end{abstract}

	\section{Introduction}
	Feature selection for high-dimensional  data is of  fundamental importance for different  applications across  various scientific disciplines \citep{tang2014feature,li2018efficient}. Grouping structure among features arises naturally in many statistical modeling problems. Common examples range from multilevel categorical features in a regression model to genetic markers from the same gene in genetic association studies. Incorporating the grouping structure information into the feature selection can take advantage of the scientifically meaningful prior knowledge, increase the feature selection accuracy and improve the interpretability of the feature selection results \citep{huang2012selective}.
	
	In this paper, we focus on group-feature selection as an approach for model interpretation and dimension reduction in both linear and nonlinear contexts. Our method can achieve stable feature selection results in a high dimensional setting when $p>n$, which is usually a challenging problem for existing methods, where $p$ is the number of features and $n$ is the number of samples.
	
	Group-feature selection has been studied from different perspectives. The group-Lasso, a generalization of the Lasso \citep{tibshirani1996regression},
	has been proposed as a mainstream approach to conduct group-wise feature selection \citep{yuan2006model}. To relax the linear constraint, \cite{meier2008group} extended the group-Lasso from linear regression to logistic regression. To speed up the computation for group-Lasso,
	\citet{yang2015fast} have further developed a more computationally tractable and efficient algorithm.
	
	However, researchers have found that the feature selection results by Lasso and group-Lasso are sensitive to the choices of tuning parameters \citep{tibshirani1996regression,su2016sparse}.
	In practice, the tuning parameter is often chosen by cross-validation (CV). But it has been reported that in the high-dimensional settings the widely adopted CV typically tends to select a large number of false features \citep{bogdan2015SLOPE}. In order to ensure the selected features are correct and replicable, several methods  have been proposed to preform feature selection while controlling  the false discovery rate (FDR)—the expected fraction of false selections among all selections.
	
	Among them, Sorted L-One Penalized Estimation (SLOPE)  \citep{bogdan2015SLOPE} and Knockoffs \citep{barber2015controlling,candes2018panning} are the state-of-the-art methods and have received the most attention.
	SLOPE was proposed to control the FDR in the classical multiple linear regression  setting. SLOPE is defined to be the solution to a penalized objective function:
	\begin{equation*}
		\argmin_b \Big\{ \frac{1}{2}\|y-Xb\|^2+J_\lambda(b) \Big\},
	\end{equation*}
	where  $J_\lambda(b)=\sum_{i=1}\lambda_i |b|_{(i)}$, with $b\in\real{p}$, $\lambda_1\geq \cdots\geq \lambda_p\geq0$, and $|b|_{(1)} \geq\cdots\geq |b|_{(p)}$ is the vector of sorted absolute values of coordinates of $b$.
	\cite{brzyski2018group} extended SLOPE method as group-SLOPE to perform group-feature selection but it is limited to linear regression.
	
	
	The notion of Knockoffs was first introduced in \citet{barber2015controlling} and improved as model-X Knockoffs by \citet{candes2018panning}. The Knockoffs variables serve as negative controls and help identify the truly important features by comparing the feature importance between original and their Knockoffs  counterpart. Originally, it is constrained to homoscedastic linear models with $n\geq p$ \citep{barber2015controlling} and later extended to a group-sparse linear regression setting by \cite{dai2016Knockoffs}.
	
	In the state-of-the-art directions of SLOPE and Knockoffs, Group-SLOPE \citep{brzyski2018group} and group-Knockoffs \citep{dai2016Knockoffs} are the only solution for group-feature selection. However, they suffer from the following limitations. (1) group-Knockoffs can only handle linear regression and are restricted to the $n>p$ setting. (2) group-SLOPE can only deal with linear regression and can not achieve robust feature selection results in a high dimensional setting when $p>n$. (3) group-SLOPE does not provide end-to-end group-wise feature selection and requires groups of features to be orthogonal to each other.
	
	To resolve all the limitations, we propose Deep-gKnock (Deep group-feature selection using Knockoffs), which combines model-X Knockoffs and Deep neural networks (DNNs) to perform model-free group-feature selection in both linear and nonlinear contexts while controlling the group-wise FDR. DNNs are a natural choice to modeling complex nonlinear relationships and performing end-to-end deep representation learning \citep{kingma2013auto} for high dimensional data. However, DNNs are often treated as black-box due to its lack of interpretability and reproducibility. Based on \citet{chen2018learning}'s work on individual level feature selection on DNNs, Deep-gKnock constructs group Knockoffs features to perform group-feature selection for DNNs.
	
\begin{figure}[!htb]
		\centering
		\includegraphics[width= .9\textwidth ]{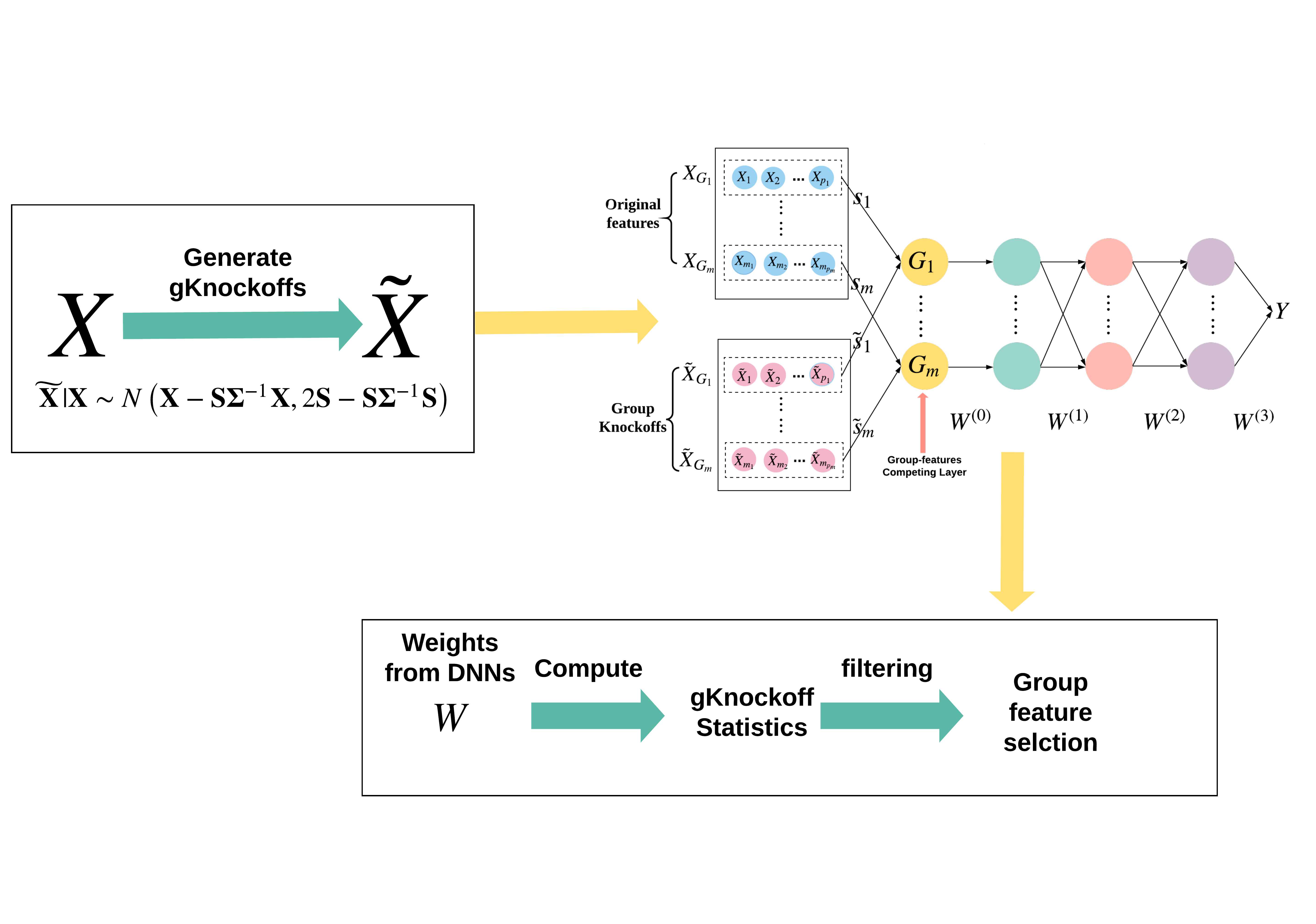}
		\caption{A graphical illustration of three steps of Deep-gKnock. This figure is best viewed in color.} \label{fig:flow}
	\end{figure}

	Figure~\ref{fig:flow} provides an overview for our
	Deep-gKnock procedure, which includes (1) generate  Group Knockoffs features; (2) incorporate original features and Group Knockoffs features into a DNN architecture to compute Knockoffs statistic; and (3) filtering out the unimportant group-feature using Knockoffs statistic. Experiment results demonstrate that our method  achieves superior power and accurate FDR control compared with state-of-the-art methods.
	
	To summarize, we make the following contributions: (1) end-to-end group-wise feature selection and deep representations for a $p>n$ setting; (2) flexible modeling framework in a DNN context with enhanced interpretability and reproducibility; (3) superior performance in terms of power and controlled group-wise false discovery rate for sythetic and real data analysis in both linear and nonlinear settings.
	
	\section{Background}
	\subsection{Problem statement}\label{sec:probelem_state}
	In our problem, we have $n$ independent and identically distributed (i.i.d) observations $\bX_i,Y_i$, where $\bX_i\in\real{p}$, $Y_i\in \real{}$, $i=1,\ldots,n$. We use $\bX_i$ to denote the feature vector and $Y_i$ to denote the scalar response variable. Denote $\chi=\{1, 2, \ldots, p\}$. We assume there exists group structure within the $p$ features, which can be partitioned into $m$ groups with group sizes $p_1,\ldots,p_m$. The index of the features in the $j$th group is denoted as $G_j$, where $\vert G_j\vert=p_j$. It satisfies that $G_j\subset\chi$ for $j=1, 2, \ldots, m$,  $ \cup_{j=1}^m G_j=\chi$ and $\cap_{j=1}^m G_j=\emptyset$. Assume that there exists a subset $\Scal_0\subset\{1,\ldots,m\}$ such that conditional on the groups of features in $\Scal_0$, the response  $Y_i$ is independent of groups of features in the complement $\Scal_0^c$. Denote $\hat{\Scal}\subset\{1,\ldots,m\}$ as the set of all the selected groups of features. Our goal is to ensure high true positive rate (TPR) defined as
	$\textrm{TPR}=\frac{\vert\hat{S}\cap S_0\vert}{\vert S_0\vert}$ while controlling the group-wise false discover rate (gFDR), which is the expected proportion of irrelevant groups among all groups of features selected and defined as
	$$\gFDR=\ep \Bigg[\frac{|\hat{\Scal}\cap \Scal_0^c|}{\max\{|\hat{\Scal}|,1\} }\Bigg].$$

	\subsection{Model-X Knockoffs framework review}
	The Knockoffs features are constructed as negative controls to help identify the truly important features by comparing the feature importance between the original and their Knockoffs counterpart. Model-X Knockoffs features are generated to perfectly mimic the arbitrary dependence structure among the original features but are conditionally independent of the response given the original features. However, model-X Knockoffs procedure \citep{candes2018panning} is only able to construct Knockoffs variables for individual feature selection. Our deep-gKnock procedure described in Section \ref{sec:group_Knockoffs} extends model-X Knockoffs procedure to generate group Knockoffs features, which allows group structure among features.
	
	For better understanding, we review the model-X Knockoffs method first. Model-X Knockoffs is designed for the individual feature selection and does not consider the grouping structure among features. So the $\Scal_0 ,\hat{\Scal}$ are defined as the indices of individual features, which are different from definitions in Section \ref{sec:probelem_state} .  Model-X Knockoffs method assume that there exists a subset $\Scal_0\subset\{1,\ldots,p\}$ such that conditional on the features in $\Scal_0$, the response  $Y_i$ is independent of features in the complement $\Scal_0^c$. We denote $\hat{\Scal}\subset\{1,\ldots,p\}$ as the set of all the selected individual features.
	
	We start this section with the model-X Knockoffs feature definition, followed by the Knockoffs feature generation process and end with the filtering process for feature selection.
	
	\begin{definition}[\citet{candes2018panning}]\label{defi.1}
		Suppose the family of random features  $\bX=(X_1,\cdots,X_p)^T$.
		Model-X Knockoffs features for $\bX$ are a new family of random features $\bXtilde=(\Xtilde_1,\cdots,\Xtilde_p)^T$ that satisfies two properties: (1) $(\bX,\bXtilde)_{\swap(\Scal)}\deq(\bX,\bXtilde)$ for any subset $\Scal\subset \{1,\cdots,p\}$, where $\swap(\Scal)$ means swapping $X_j$ and $\Xtilde_j$ for each $j\in\Scal$ and $\deq$ denotes equal in distribution, and (2) $\bXtilde \bot Y |\bX$, i.e., $\bXtilde$ is independent of response $Y$ given feature $\bX$.
	\end{definition}
	From this definition, we can see that model-X Knockoffs feature $\Xtilde_j$'s  mimic dependency structure among the original features $X_j$'s and are independent of the response $Y$ given $X_j$'s. By comparing the original features $\bX$ with the Knockoffs features $\bXtilde$, FDR can be controlled at target level $q$. When $X\sim N(0,\bSigma)$ with the covariance matrix $\bSigma\in\real{p\times p}$ , we can construct the model-X Knockoffs features $\bXtilde$ characterized in Definition \ref{defi.1} as
	\begin{equation}\label{eqn_distribution1}
	\bXtilde|\bX \sim N(\bX-\diag\{\bs\}\bSigma^{-1}\bX, 2 \diag\{\bs\}-\diag\{\bs\}\bSigma^{-1}\diag\{\bs\}).
	\end{equation}
	Here $\diag\{\bs\}$ with all components of $\bs\in\real{p}$ being positive is a diagonal matrix with requirement that the conditional covariance matrix in Equation \ref{eqn_distribution1} is positive definite. Following the above Knockoffs construction, the joint distribution of the original features and the model-X Knockoffs features is
	
	\begin{equation}\label{eqn_distribution2}
	(\bXtilde,\bX) \sim N\Bigg(\begin{pmatrix} 0\\0\end{pmatrix} ,
	\begin{pmatrix}
	\bSigma & \bSigma-\diag\{\bs\}\\
	\bSigma-\diag\{\bs\}  & \bSigma
	\end{pmatrix}
	\Bigg).
	\end{equation}
	
	To ensure high power in distinguishing $\bX$ and $\bXtilde$, it is desired that the constructed Knockoffs features $\bXtilde$ deviate from the original features $\bX$ while maintaining the same correlation structure as $\bX$. This indicates larger components of $\bs$ are preferred since $\Cov(\bX, \bXtilde) = \bSigma-\diag\{\bs\}$. In a setting where the features are normalized, i.e. $\bSigma_{jj}=1$ for all $j$, we would like to have $\Cor(X_j,\Xtilde_k)=1-s_j$ as close to zero as possible. One way to choose $\bs$ is the equicorrelated construction \citep{barber2016knockoff}, which uses
	$$s_j^\EQ=2\lambda_{\min}(\bSigma)\wedge 1\textrm{ for all } j.$$
	
	Then we define the Knockoffs statistic $W_j$ for  each feature $X_j$, $j\in \{1,\dots,p\}$, which is used in the filtering process to perform feature selection. A large positive value of $W_j$ provides evidence that $X_j$ is important. This statistic depends on $\bX,\bXtilde$ and $Y$, i.e. $W_j=w_j((\bX,\bXtilde),Y)$  for some function $w_j$. This function $w_j$ must satisfy the following flip-sign property:
	\begin{equation}\label{eqn:flip}
	w_{j}\left([\bX, \bXtilde]_{\operatorname{swap}(S)}, y\right)=\left\{\begin{array}{ll}{w_{j}([\bX, \bXtilde], y),} & {j \notin S} \\ {-w_{j}([\bX, \bXtilde], y),} & {j \in S}\end{array}\right.
	\end{equation}	
	\cite{candes2018panning} construct the Knockoffs statistic by performing Lasso on the original features $\bX$ augmented with Knockoffs $\bXtilde$	
	$$
	\min _{b \in \mathbb{R}^{2 p}} \frac{1}{2}\|y-[\bX, \bXtilde] b\|_{2}^{2}+\lambda\|b\|_{1},
	$$	
	which provides Lasso coefficients $b_1,\ldots,b_{2p}$. The statistic $W_j$ is set to be the Lasso coefficient difference given by	
	$$W_j=|b_j|-|b_{j+p}|$$	
	After obtaining the Knockoffs statistic satisfying \eqref{eqn:flip}, Theorem~\ref{theorem:filter} from \cite{candes2018panning} provides feature selection procedure with controlled FDR.
	\begin{theorem}[\cite{candes2018panning}]\label{theorem:filter}
		Let $q\in[0,1]$. Given statistic, $W_1,\ldots,W_p$ satifying \eqref{eqn:flip}, let
		$$
		\tau=\min \left\{t>0 : \frac{1+\#\left\{j : W_{j} \leq-t\right\}}{\#\left\{j : W_{j} \geq t\right\}} \leq q\right\}.
		$$
		Then the procedure selecting the features $\hat{S}=\{l:W_j\geq \tau\}$, controls the FDR at level $q$.
	\end{theorem}	
	\section{Deep group-feature selection using Knockoffs}\label{sec:group_Knockoffs}
	\subsection{Constructing Group Knockoffs features}
	The original Knockoffs construction \citep{candes2018panning} does not take group structure among different features into account and requires stronger constraints. When there exists high correlation between features $X_j$ and $\Xtilde_j$, \citet{candes2018panning}'s method requires that the values of $\bs$ to be extremely small in order to ensure the covariance matrix in Equation~\eqref{eqn_distribution2} is positive semi-definite. However, smaller values of $\bs$ will fail to detect the difference between $\bX$ and $\bXtilde$, which will lead to a decrease in the power of detecting the true positive features.
	In a group-sparse setting, we relax this requirement by proposing our Group Knockoffs features in Definition~\ref{defi.2} to increase the power.
	
	\begin{definition}[Group Knockoffs features]\label{defi.2}
		Suppose the family of random features  $\bX=(X_1,\cdots,X_p)^T$ has group structure, where the $p$ features are partitioned into $m$ groups, $G_1,\ldots,G_m \subset\chi=\{1,\ldots,p\}$, with group sizes $p_1,\ldots,p_m$, $ \cup_{j=1}^m G_j=\chi$ and $\cap_{j=1}^m G_j=\emptyset$.
		Group Knockoffs features for $\bX=(X_1,\cdots,X_p)^T$ are a new family of random features $\bXtilde=(\Xtilde_1,\cdots,\Xtilde_p)^T$ that satisfies two properties: (1) $(\bX,\bXtilde)_{\swap(\Scal)}\deq(\bX,\bXtilde)$ for any subset $\Scal\subset \{1,\cdots,m\}$, where $\swap(\Scal)$ means swapping $\bX_{G_j}$ and $\bXtilde_{G_j}$ for each $j\in\Scal$ and $\deq$ denotes equal in distribution, and (2) $\bXtilde \bot Y |\bX$, i.e., $\bXtilde$ is independent of the response $Y$ given feature $\bX$.
	\end{definition}
	We see from this definition, that the Group Knockoffs features $\Xtilde_j$'s  mimic the group-wise dependency structure among the original features $X_j$'s and are independent of the response $Y$ given $X_j$'s. When $X\sim N(0,\bSigma)$, the joint distribution obeying Definition \ref{defi.2} is
	\begin{equation}\label{eqn_distribution3}
	(\bXtilde,\bX) \sim N\Bigg(\begin{pmatrix} 0\\0\end{pmatrix} ,
	\begin{pmatrix}
	\bSigma & \bSigma-\bS\\
	\bSigma-\bS  & \bSigma
	\end{pmatrix}
	\Bigg)   .
	\end{equation}
	where $\bS=\diag(\bS_1,\ldots,\bS_m)\prec 2\bSigma$ is a group-block-diagonal matrix. Here we use $\bA \prec \bB$ to denote $\bB-\bA$ is positive definite.
	
	We construct the Group Knockoffs features by sampling the Knockoffs  vector $\bXtilde$ from the conditional distribution
	\begin{equation}\label{eqn_distribution4}
	\bXtilde|\bX \sim N(\bX-\bS\bSigma^{-1}\bX, 2 \bS-\bS\bSigma^{-1}\bS).
	\end{equation}
	
	Follow \cite{dai2016Knockoffs}, the group-block-diagonal matrix $\bS=\diag(\bS_1,\ldots,\bS_m)$ satisfying $\bS\prec 2\bSigma$ can be constructed with
	
	\vspace{-0.2cm}
	$$\bS_i=\eta \bSigma_{G_i,G_i},\ \textrm{where }\eta=2\lambda_{\min}(\bD\bSigma\bD)\wedge 1,\ \bD=\diag\{\bSigma_{G_1G_1}^{-1/2} ,\ldots,\bSigma_{G_mG_m}^{-1/2}\}.$$
	
	\subsection{Deep neural networks for Group Knockoffs features}
\begin{figure}[ht]
		\centering
		\includegraphics[width= .9\textwidth ]{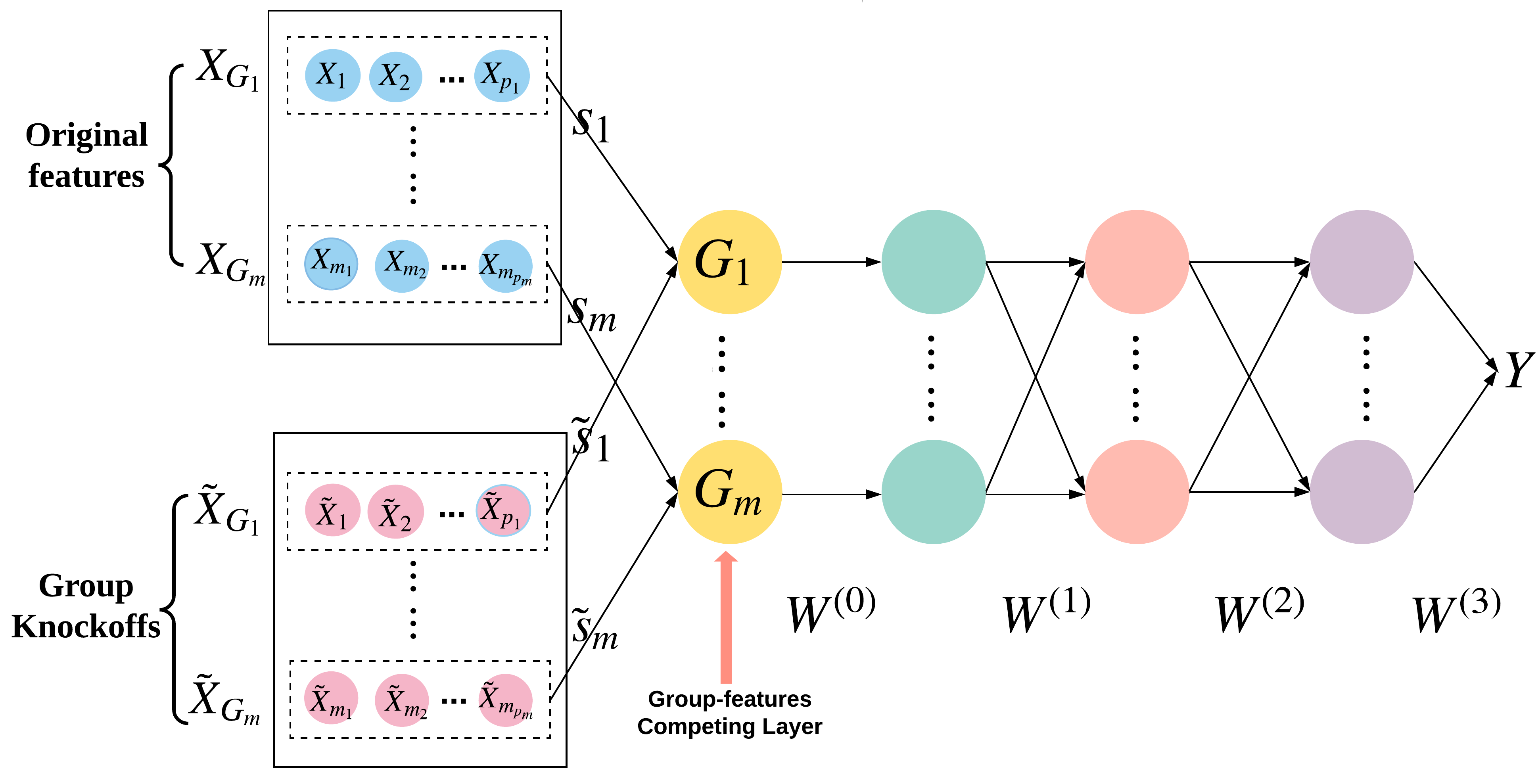}
		\caption{A graphical demonstration of the DNN structure for Deep-gKnock. This figure is best viewed in color.} \label{fig:DNN}
	\end{figure}

	Once the Group Knockoffs features are constructed, following  similar idea in DeepPINK \citep{lu2018deeppink}, we feed them into a new DNN structure to obtain gKnock statistic. The structure of the network is shown in Figure \ref{fig:DNN}.
	
	In the first layer, we feed $(\bX,\bXtilde)$ into a Group-feature Competing Layer containing $m$ filters, $G_1,\ldots,G_m$. The $j$th filter $G_j$ connects group-feature $\bX_{G_j}$ and its Knockoffs counterpart  $\bXtilde_{G_j}$. We use a linear activation function in this layer to encourage the competition between group-feature and its Knockoffs counterpart. Intuitively, if the group-feature $\bX_{G_j}$ is important, we expect the magnitude of $S_j$ to be much larger than $\tilde{S}_j$, and if the the group-feature $\bX_{G_j}$ is not important, we expect the magnitude of $S_j$ and  $\tilde{S}_j$ to be similar.
	
	We then feed the output of the Group-feature Competing Layer into  a fully connected multilayer perceptron (MLP) to learn a non-linear mapping to the response $Y$.  We use  $W^{(0)}\in\real{m\times1}$ to denote the weight vector connecting the Group-features Competing Layer to the MLP.
	The MLP has two hidden layers, each containing $m$ neurons, and ReLU activation and $L_1$-regularization are used, as shown in Figure \ref{fig:DNN}. We use $W^{(1)}\in\real{m\times m}$ to denote the weight matrix connecting the input vector to the first hidden layer. Similarly, we use $W^{(2)}\in\real{m\times m}$ as the weight matrix connecting two hidden layers and  $W^{(3)}\in\real{m\times 1}$ as the weight matrix connecting second hidden layer to the output $Y$.

	\subsection{gKnock statistic}
	After the DNN is trained, we compute gKnock statistic based on the weights to evaluate the importance of group-feature. Firstly, we use $\bz=(\|S_1\|_2^2/p_1,\ldots,\|S_m\|_2^2/p_m)^T$ and $\bztilde=(\|\tilde{S}_1\|_2^2/p_1,\ldots,\|\tilde{S}_m\|_2^2/p_m)^T$ to represent the relative importance between $\bX_{G_j}$ and $\bXtilde_{G_j}$ $j=1,\ldots,m$.  Secondly, we assess the relative importance of the $j$th group-feature among all $m$ group-feature by  $\bw=W^{(0)}\circ (W^{(1)}W^{(2)}W^{(3)})\in\real{m\times 1}$, where $\circ$ denotes the Schur (entrywise) matrix product. Thirdly, the importance measures for $\bX_{G_j}$ and $\bXtilde_{G_j}$ are provided by
	\begin{equation*}\label{eqn:gKnock-stat}
		Z_j= \|S_j\|_2^2 \times w_j\quad \textrm{and}\quad  	\tilde{Z}_j= \|\tilde{S}_j\|_2^2 \times \tilde{w}_j.
	\end{equation*}
	Finally, we define the gKnock statistic as
	$$W_j=Z_j^2-\tilde{Z}_j^2,\quad j=1,\ldots,m$$
	and the same filtering process as Theorem \ref{theorem:filter} is applied to the $W_j$'s to select group-feature.

	\section{Simulation studies}\label{sec:simulation}
	We evaluate the performance of our method both in Gaussian linear regression model \eqref{eqn.lm} and Single-Index model \eqref{eqn.SIM}.
	\begin{equation}\label{eqn.lm}
	Y_i=\bX_i^T\bbeta+\eps_i,\qquad i=1,\ldots,n,
	\end{equation}
	\begin{equation}\label{eqn.SIM}
	Y_i=g(\bX_i^T\bbeta)+\eps_i,\qquad i=1,\ldots,n,
	\end{equation}
	where $Y_i\in\real{}$ is the $i$th response, $\bX_i\in\real{p}$ is the feature vector of the $i$th observation, $\bbeta\in\real{p}$ is the coefficient vector, $\epsilon_i\in\real{p}$ is the noise of $i$th observation, and $g$ is some unknown link function.

	To generate the sythetic data, we set the number of feature $p=1000$ and the number of groups $g=100$ with the number of features per group as $p_i=10$. The true regression coefficient vector $\bbeta_0\in\real{p}$ is group sparse with $k=20$ groups of nonzero signals,  and the nonzero coefficients are randomly chosen from $\{\pm 1.5\}$. We draw $X_i$ independently from a multivariate normal distribution with mean $\bm{0}$ and covariance matrix $\bSigma$, with diagonal entries $\Sigma_{ii}=1$, within-group correlations $\Sigma_{ij}=\rho$ for $i\neq j$ in the same group, between-group correlations $\Sigma_{ij}=\gamma\rho$ for $i$, $j$ in the different groups. The error $\epsilon_i$ are i.i.d. from standard normal distribution.  The true link function is $g(x)=(x/20)^3+4(x/20)^2$.

	In our default setting, we set $n=1000$, $\rho=\gamma=0$.	To study the effects of sample size, between-group correlation and within-group correlation, we vary one setting and keep the others remain at their default level in each experiment.
	\begin{itemize}\setlength\itemsep{0.2cm}
		\item Sample size: we vary the number of observations from  500,750,1000,1250 to 1500.
		\item Group correlation: we fix the within-group correlation $\rho=0.5$, and set the between-group correlation to be $\gamma\rho$, with $\gamma\in\{0,0.2,\ldots,0.8\}$.
		\item Within-group correlation: we vary within-group correlation with $\rho\in\{0,0.2,\ldots,0.8\}$ and fix $\gamma=0.4$.
	\end{itemize}
	We compare the performance  of Deep-gKnock with group-SLOPE  available in the R package grpSLOPE \citep{grpSLOPE}. 	For each setting, we run each experiment for 100 replications and set the target gFDR level $q=0.2$. The empirical gFDR and power are reported in Table \ref{tab1} \& \ref{tab2}.
	
	In the linear model setting shown in Table~\ref{tab1}, group-SLOPE fails to control the gFDR at the target level $\textrm{gFDR}=0.2$ in each of the following three situations: (1) $p>n$; (2) between-group correlation $\gamma$ is large; (3) within group correlation $\rho$ is large. In contrast,  Deep-gKnock  can precisely control the gFDR in all settings.
	
	In the single-index model setting shown in Table~\ref{tab2}, Deep-gKnock achieves higher power and consistently controls gFDR in all settings, which demonstrates the advantages of our Deep-gKnock by using DNN to model the non-linear relationship between features and the response.
	\begin{table}[h!]
		\setlength{\tabcolsep}{3pt}
		{\footnotesize \centering
			\caption{Simulation results for linear model.}\label{tab1}
			\begin{tabular}{lllllllllllllll}
				\cmidrule(lr){1-5} \cmidrule(lr){6-10} \cmidrule(lr){11-15}
				\multicolumn{5}{c}{Varying Sample size $n$}& \multicolumn{5}{c}{Varying Between-group correlation $\gamma$}   & \multicolumn{5}{c}{Varying Within-group correlation $\rho$} \\
				\cmidrule(lr){1-5} \cmidrule(lr){6-10} \cmidrule(lr){11-15}
				\multicolumn{1}{c}{\multirow{2}{*}{$n$}} & \multicolumn{2}{c}{Deep-gKnock}                     & \multicolumn{2}{c}{group-SLOPE}                                                   &\multicolumn{1}{c}{\multirow{2}{*}{$\gamma$}} & \multicolumn{2}{c}{Deep-gKnock}                     & \multicolumn{2}{c}{group-SLOPE}                                             & \multicolumn{1}{c}{\multirow{2}{*}{$\rho$}}& \multicolumn{2}{c}{Deep-gKnock}                     & \multicolumn{2}{c}{group-SLOPE}                                            \\
				\cmidrule(lr){2-3} \cmidrule(lr){4-5}\cmidrule(lr){7-8} \cmidrule(lr){9-10}\cmidrule(lr){12-13} \cmidrule(lr){14-15}
				\multicolumn{1}{c}{}                   & \multicolumn{1}{c}{gFDR} & \multicolumn{1}{c}{Power} & \multicolumn{1}{c}{gFDR} & \multicolumn{1}{c}{Power}               & & \multicolumn{1}{c}{gFDR} & \multicolumn{1}{c}{Power} & \multicolumn{1}{c}{gFDR} & \multicolumn{1}{c}{Power} &                      & \multicolumn{1}{c}{gFDR} & \multicolumn{1}{c}{Power} & \multicolumn{1}{c}{gFDR} & \multicolumn{1}{c}{Power} \\ 	
				\cmidrule(lr){1-5} \cmidrule(lr){6-10} \cmidrule(lr){11-15}
				500 & \textbf{0.19} & 0.98 & 0.36 & 0.73 & 0.00 & \textbf{0.18} & 0.98 & 0.20 & 1.00 & 0.00 & \textbf{0.17} & 1.00 & 0.21 & 1.00 \\
				750 & \textbf{0.21} & 0.99 & 0.30 & 0.99 & 0.20 & \textbf{0.18} & 0.99 & 0.23 & 1.00 & 0.20 & \textbf{0.19} & 1.00 & 0.22 & 1.00 \\
				1000 & \textbf{0.20} & 0.99 & 0.21 & 1.00 & 0.40 & \textbf{0.20} & 0.99 & 0.26 & 1.00 & 0.40 & \textbf{0.14} & 1.00 & 0.24 & 1.00 \\
				1250 & \textbf{0.23} & 0.99 & 0.17 & 1.00 & 0.60 & \textbf{0.17} & 0.99 & 0.30 & 1.00 & 0.60 & \textbf{0.14} & 1.00 & 0.27 & 1.00 \\
				1500 & \textbf{0.21} & 0.99 & 0.15 & 1.00 & 0.80 & \textbf{0.18} & 0.99 & 0.40 & 1.00 & 0.80 & \textbf{0.11} & 0.95 & 0.30 & 1.00 \\
				\cmidrule(lr){1-5} \cmidrule(lr){6-10} \cmidrule(lr){11-15}
			\end{tabular}
			
		}
	\end{table}

	\begin{table}[h!]
		\setlength{\tabcolsep}{3pt}
		{\footnotesize \centering
			\caption{Simulation results for Single-Index model.}\label{tab2}
			\begin{tabular}{lllllllllllllll}
				\cmidrule(lr){1-5} \cmidrule(lr){6-10} \cmidrule(lr){11-15}
				\multicolumn{5}{c}{Varying Sample size $n$}& \multicolumn{5}{c}{Varying Between-group correlation $\gamma$}   & \multicolumn{5}{c}{Varying Within-group correlation $\rho$} \\
				\cmidrule(lr){1-5} \cmidrule(lr){6-10} \cmidrule(lr){11-15}
				\multicolumn{1}{c}{\multirow{2}{*}{$n$}} & \multicolumn{2}{c}{Deep-gKnock}                     & \multicolumn{2}{c}{group-SLOPE}                                                   &\multicolumn{1}{c}{\multirow{2}{*}{$\gamma$}} & \multicolumn{2}{c}{Deep-gKnock}                     & \multicolumn{2}{c}{group-SLOPE}                                             & \multicolumn{1}{c}{\multirow{2}{*}{$\rho$}}& \multicolumn{2}{c}{Deep-gKnock}                     & \multicolumn{2}{c}{group-SLOPE}                                            \\
				\cmidrule(lr){2-3} \cmidrule(lr){4-5}\cmidrule(lr){7-8} \cmidrule(lr){9-10}\cmidrule(lr){12-13} \cmidrule(lr){14-15}
				\multicolumn{1}{c}{}                   & \multicolumn{1}{c}{gFDR} & \multicolumn{1}{c}{Power} & \multicolumn{1}{c}{gFDR} & \multicolumn{1}{c}{Power}               & & \multicolumn{1}{c}{gFDR} & \multicolumn{1}{c}{Power} & \multicolumn{1}{c}{gFDR} & \multicolumn{1}{c}{Power} &                      & \multicolumn{1}{c}{gFDR} & \multicolumn{1}{c}{Power} & \multicolumn{1}{c}{gFDR} & \multicolumn{1}{c}{Power} \\ 	
				\cmidrule(lr){1-5} \cmidrule(lr){6-10} \cmidrule(lr){11-15}
				500 & 0.22 & \textbf{0.71} & 0.08 & 0.03 & 0.00 & 0.14 & \textbf{0.53} & 0.12 & 0.17 & 0.00 & 0.20 & \textbf{0.78} & 0.12 & 0.18 \\
				750 & 0.18 & \textbf{0.72} & 0.14 & 0.15 & 0.20 & 0.19 & \textbf{0.74} & 0.30 & 0.28 & 0.20 & 0.25 & \textbf{0.79} & 0.31 & 0.31 \\
				1000 & 0.18 & \textbf{0.72} & 0.12 & 0.21 & 0.40 & 0.20 & \textbf{0.82} & 0.46 & 0.35 & 0.40 & 0.17 & \textbf{0.83} & 0.42 & 0.34 \\
				1250 & 0.18 & \textbf{0.73} & 0.12 & 0.32 & 0.60 & 0.21 & \textbf{0.88} & 0.52 & 0.40 & 0.60 & 0.17 & \textbf{0.88} & 0.48 & 0.35 \\
				1500 & 0.19 & \textbf{0.75} & 0.14 & 0.45 & 0.80 & 0.19 & \textbf{0.86} & 0.57 & 0.43 & 0.80 & 0.17 & \textbf{0.94} & 0.53 & 0.34 \\
				\cmidrule(lr){1-5} \cmidrule(lr){6-10} \cmidrule(lr){11-15}
			\end{tabular}
		}
	\end{table}
	\vspace{-.6cm}	
	\section{Real data analysis}
	In addition to the simulation studies presented in Section~\ref{sec:simulation}, we also demonstrate the performance of Deep-gKnock on two real data sets. The gFDR level is set to $q=0.2$.
	
	\subsection{Application to prostate cancer data}
	The prostate cancer data contains  clinical measurements for 97 male patients who were about to receive a radical prostatectomy. It was analyzed in \cite{hastie2013elements} to study  the correlation between the response $Y$, the level of prostate-specific antigen (lpsa) and other eight features. The features are  log cancer volume (lcavol), log prostate weight (lweight), age, log of the amount of benign prostatic hyperplasia (lbph), seminal vesicle invasion (svi), log of capsular penetration (lcp), Gleason score (gleason), and percent of Gleason scores 4 or 5 (pgg45).
	
	For the categorical variable svi with two levels, we coded it by one dummy variable and treated it as one group. For each continuous variable, we used five B-Spline basis functions to represent its effect and treated those five basis functions as a group. This provides us eight groups with a total of 36 features. We summarize the group-feature selection results in Table~\ref{tab_prostate}. The features selected by Deep-gKnock are the same as using Lasso in \citet{hastie2013elements}.
	
	\begin{table}[h!]
		\begin{center}
			\caption{Group-feature selection results for prostate cancer data}\label{tab_prostate}		
			\begin{tabular} {c|c }
				\hline
				Method & group-feature selected   \\ \hline
				group-SLOPE &  lcavol, lweight, svi, gleason\\
				Deep-gKnock & lcavol, lweight \\
				\hline
			\end{tabular}
		\end{center}
	\end{table}
	\vspace{-.2cm}	
	\subsection{Application to yeast cell cycle data}
	We apply Deep-gKonck to the task of identifying the important transcription factors (TFs), which are related to regulation of the cell cycle.   TFs belong to a class of proteins called    binding proteins, and control the rate at which DNA is transcribed into mRNA. We utilize a yeast cell cycle data set from \cite{spellman1998comprehensive}  and \cite{lee2002transcriptional}. The response $Y$  is the messenger ribonucleic acid (mRNA)  levels   on $n=542$ genes, and are measured at 28 minutes during a cell cycle. The features $\bX$ is the  measurements of binding information of $p=106$ TFs .  Out of the 106 TFs, 21 TFs are known and experimentally confirmed cell cycle related TFs \citep{wang2007group}.

\begin{figure}[ht]
		\centering
		\includegraphics[width= .5\textwidth ]{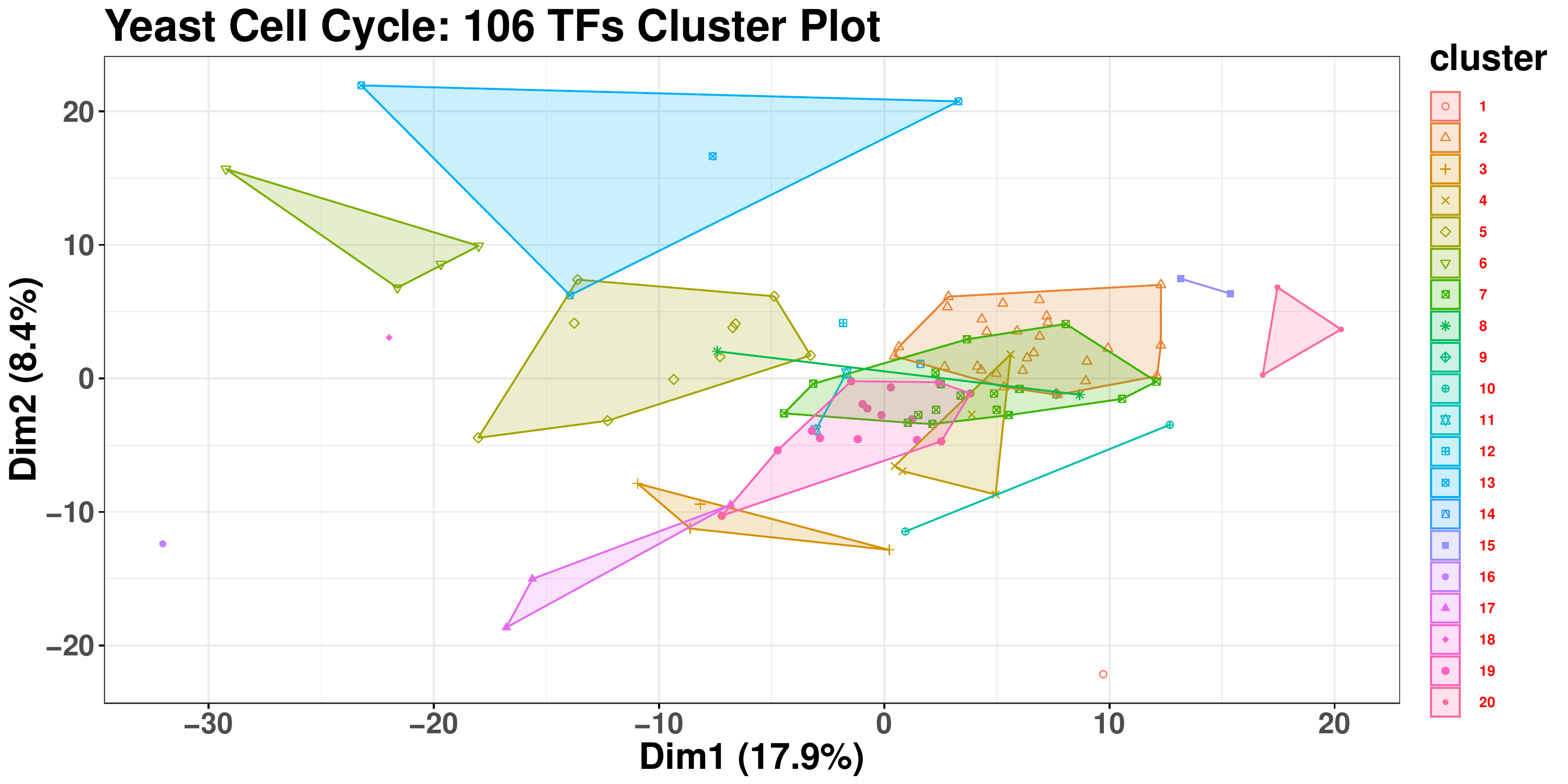}
		\caption{Cluster plot for 106 TFs in Yeast Cell Cycle data} \label{fig:yeast-cluster}
	\end{figure}

	It has been studied that groups of TFs function in a coordinated fashion to direct cell division, growth and death \citep{latchman1997transcription}. Following \cite{ma2007supervised}, we use the K-means method to cluster the 106 TFs, and determine the optimal number of clusters using the Gap statistic \citep{tibshirani1999clustering}. The Gap statistic suggests the 106 TFs can be clustered into 20 groups. To visulize the clustering results, we use Principal Component Analysis (PCA) algorithm to reduce the dimensionality to its first two principal components, which results in a scatter plot of data points colored by their cluster labels in Figure \ref{fig:yeast-cluster}.  One of the clusters contains four TFs and all of them are experimentally verified.
	\vspace{-.1cm}	
	Group-SLOPE identified 7 groups which contains 41 TFs. including 12 confirmed TFs. Deep-gKnock  identified  5 groups which contains 26 TFs including  11 confirmed TFs. To demonstrate the selection performance, following \cite{zhu2019envelope-spls}, we also compute the probability of containing at least $q$ confirmed TFs from a $s$ randomly chosen TFs from a hypergeometric distribution in Table \ref{tab_prob}. We included the results for the Lasso in Table \ref{tab_prob} as a benchmark. Smaller probability values suggest better feature selection performance. The small probability of Deep-gKnock suggests that the large number of confirmed TFs selected is not due to chance. Deep-gKnock also outperforms group-SLOPE.
	\begin{table}[h!]
		\begin{center}
			\caption{Probability of containing at least  $q$ confirmed TFs out of 85 unconfirmed and 21 confirmed TFs in a random draw of $s$ TFs.}\label{tab_prob}		
			\begin{tabular} {c|c c c}
				\hline
				Method & $s$ & $q$ & $P(Q\geq q)$   \\ \hline
				Lasso & $100$ & $21$ & $0.256$\\
				group-SLOPE & $41$ & $12$ & $0.04673$\\
				Deep-gKnock & $26$ & $11$ & $0.00192$\\
				\hline
			\end{tabular}
		\end{center}
	\end{table}
	\vspace{-1cm}
	\section{Conclusion}
	We have introduced a novel group-feature selection method Deep-gKnock combining Knockoffs with DNNs. It provides
	an end-to-end group-wise feature selection with controlled gFDR for high dimensional data. With the flexibility of DNN, we also provide deep representations with enhanced interpretability and reproducibility. Both synthetic and real data analysis is provided to demonstrate that Deep-gKnock can achieve superior power and accurate gFDR control compared with state-of-the-art methods. Moreover, Deep-gKnock achieves scientifically meaningful group-feature selection results for real data sets.
	
	\bibliographystyle{natbib}
	
	\bibliography{Deep-gKnock}

\end{document}